\documentclass[10pt,twocolumn,letterpaper]{article}

\usepackage[pagenumbers]{cvpr} 

\usepackage{graphicx}
\usepackage{amsmath}
\usepackage{amssymb}
\usepackage{booktabs}

\usepackage{microtype}

\usepackage[pagebackref,breaklinks,colorlinks]{hyperref}

\usepackage[capitalize]{cleveref}
\crefname{section}{Sec.}{Secs.}
\Crefname{section}{Section}{Sections}
\Crefname{table}{Table}{Tables}
\crefname{table}{Tab.}{Tabs.}

\def\cvprPaperID{10} 
\def\confName{CVPR}
\def\confYear{2022}

\begin{document}

\title{Segmentation Enhanced Lameness Detection in Dairy Cows from RGB and Depth Video}


\author{Eric Arazo$^1$, Robin Aly$^2$, Kevin McGuinness$^1$\\
$^1$Insight Centre for Data Analytics, Dublin City University (DCU), Ireland\\
$^2$Nedap N.V., The Netherlands\\
{\tt\small eric.arazo@insight-centre.org, robin.aly@nedap.com,  kevin.mcguinness@insight-centre.org}
}

\maketitle

\begin{abstract}
    Cow lameness is a severe condition that affects the life cycle and life quality of dairy cows and results in considerable economic losses. Early lameness detection helps farmers address illnesses early and avoid negative effects caused by the degeneration of cows' condition. We collected a dataset of short clips of cows passing through a hallway exiting a milking station and annotated the degree of lameness of the cows. This paper explores the resulting dataset and provides a detailed description of the data collection process. Additionally, we proposed a lameness detection method that leverages pre-trained neural networks to extract discriminative features from videos and assign a binary score to each cow indicating its condition: ``healthy" or ``lame." We improve this approach by forcing the model to focus on the structure of the cow, which we achieve by substituting the RGB videos with binary segmentation masks predicted with a trained segmentation model. This work aims to encourage research and provide insights into the applicability of computer vision models for cow lameness detection on farms.
\end{abstract}

\section{Introduction}
\label{sec:intro}
Detection of lameness in cows is one of the main focuses of health and welfare monitoring in dairy farming. Lameness is a condition characterized by restricted mobility and difficulties in the gait due to lesions in the foot or limbs: it affects the cow reproductive cycle, milk yield, and life quality; and it is responsible for substantial economic losses~\cite{2021_sensors_review, 2019_elsevier_review}. Consequently, early cow lameness detection is of utmost importance for farmers to address the illness and avoid the degeneration into severe cases resulting in further welfare and economical complications. 
To facilitate early detection, veterinaries visually assign to each cow a locomotion degree that asses the severity of each case.
This scoring system, further discussed in
Section~\ref{sec:dataset_and_task}, presents some initial challenges for building an automated system to detect lameness, these include the need of expert knowledge for the annotation of datasets, and the subjectivity of the annotations, which often results in disagreement between experts.

In this scenario where visual features are so relevant, computer vision techniques provide a promising solution to automate this process: these are non-intrusive systems that leverage visual features to solve a particular task. Computer vision has recently shown outstanding results in a large variety of tasks such as medical imaging~\cite{2021_Wiley_review}, herbage mass estimation~\cite{2021_ICCVw_grass}, and person re-identification~\cite{2022_WACVw_tempConst}. While this success requires large amounts of labeled data, recent works apply transfer learning techniques to bypass the costly data collection process~\cite{2021_elsevier_eeg, 2021_CVPR_transfer}: they leverage large generic datasets 
and small amounts of data in the domain of interest to achieve outstanding results. This is particularly important in
specific domains, where 
high-quality labels are scarcer, such as medical imaging or lameness detection in cows.

Detecting lameness using computer vision requires answering several design questions.
Among the most important are: 1) camera position, where the most promising settings are either from the side, clearly capturing all involved body parts from one side, or from the top, indirectly showing the effects on all four limbs, 2) the input data that is used for classification, and 3) the classes used for classification due to the bias towards healthy cows. We evaluate every combination of answers, based on classification effectiveness. 

This paper answers the aforementioned research questions. We collect a proprietary dataset of cows passing through an ally. We record them both from the side and the top and a trained annotator decides about their lameness status. Using this dataset, we investigate the effects of changes in camera position and employed features using a state-of-the-art classifier for video sequences.

\section{Literature review}
\label{sec:literature_review}
This section is a reference for the reader to further explore lameness detection rather than a thorough review of the literature. Since the wide success of computer vision approaches in practical applications, several works have tried to automate lameness detection and animal farming in general. Kang et al.~\cite{2021_sensors_review} and Li et al.~\cite{2021_MDPI_review} provide a thorough review of recent research that leverages computer vision-based approaches for lameness detection and animal farming. In this same direction, Alsaaod et al.~\cite{2019_elsevier_review} review the literature in a broader sense and divide the approaches into kinematic methods (that explore the motion of the subjects, which include image processing techniques), kinetic methods (that explore the force distribution generated by the subjects, which include accelerometers and weighing systems), and indirect methods (that combine milking and activity sensors).

The main advantage of computer vision approaches for cow lameness detection~\cite{2020_elsevier_YOLOlameness, 2020_elsevier_cowStructure, 2021_AAAI_dl4lameness} is their low intrusiveness and low operational costs. These approaches do not interfere with cow movement and collect data from several subjects with a single camera.
Wu et al.~\cite{2020_elsevier_YOLOlameness} for example, propose to use leg coordinates extracted from cow videos with an object detection model (YOLOv3~\cite{2018_arxiv_yolov3}) to build a feature that encodes relative step size between the rear and front legs of the cows. Then, this feature is used in a recurrent neural network (concretely an LSTM network) to predict lameness in cows.
In this work, we describe the data collection pipeline, propose a
method for lameness detection, and improve the method by forcing the model to focus on the cow structure.

\begin{table}
\centering{}
\caption{
Annotated samples across the five locomotion scores.}\label{tab:t1}
\vskip -0.1in
\resizebox{0.3\textwidth}{!}{%
\begin{tabular}{lcccccc}
\toprule 
\multicolumn{2}{c}{ } & \multicolumn{5}{c}{Locomotion score} \tabularnewline
\midrule 
{View} & {Total} & 1 & 2 & 3  & 4  & 5 \tabularnewline
\midrule 
Side  & 869 & 607 & 240 & 17 & 3 & 2 \tabularnewline
Top & 864 & 603 & 239 & 17 & 3 & 2  \tabularnewline
\bottomrule
\end{tabular}
}
\vskip -0.15in
\end{table}
%
\section{Dataset and task}
The dataset consists of RGB and depth video fragments of the right side and top views of cows walking through a corridor after the milking station. We recorded the fragments in an installation on one farm in the Netherlands during July and August 2020. For both views, we used Intel Realsense D435 cameras connected to a Raspberry Pi 4b. We used a frame rate of 30 fps and a resolution of 640~$\times$~480 pixels. To identify passing cows, we used a Nedap walkthrough antenna with ISO FDX identification. The cows were already equipped with a Nedap Smarttag Neck with RFID, which we used to identify cows when they were passing the antenna. When the antenna identified a cow, we recorded the RGB and depth data seven seconds before and three seconds after the identification. These durations were empirically defined. The depth videos are converted to RGB video using the HUE color space as in~\cite{image_compression}. We manually filtered the fragments to only contain valid recordings, filtering out those that e.g. contain multiple cows or show the farmer.

The final dataset comprises 5164 videos. From these, 869 are side view and 864 are top view, corresponding to 116 unique cows, which were manually annotated. There is an average of eight annotated videos per cow. 
\label{sec:dataset_and_task}

To measure the lameness of cows, we used the locomotion score: a score of one is given to healthy cows, and scores two to five to cows with increasing degrees of lameness, five being the most severe. The locomotion score is assigned based on spine curvature of the cow, lame cows have a more pronounced curvature, and distance between front and back legs, shorter distances are expected in lamer cows. Table~\ref{tab:t1} shows how the fragments are distributed across locomotion scores.  These locomotion scores were annotated to each fragment by an annotator who was previously trained by a veterinarian. The assessment was done based on the recorded video, while we ensured a high agreement between this mode and annotations from observing the animals in real life. 
 
\section{Cow lameness detection}
\label{sec:cow_lameness_detection}
This section describes the preprocessing of the dataset, the segmentation stage including the annotation and training process, and the structure of the proposed model.

\subsection{Preprocessing and dataset split}
To create the train and validation splits, we used the unique identification numbers of each cow to avoid crossovers between train and validation. In other words, we make sure that individual cows appear either in the train or in the validation split, not in both. Then we aimed at a balanced validation set and selected 10 cows with all the visits labeled as one (``healthy") and 10 with at least one visit labeled from 2 to 5 (``lame"). Table~\ref{tab:t2} shows the resulting distribution of samples across locomotion scores and the distribution of samples once the samples are grouped into a binary setup: ``healthy" and ``lame." Note that the most severe cases do not have samples in the validation set, since the most challenging scenario is to distinguish between classes one and two, we assume this is not a problem and discuss the class imbalance of the dataset in Section~\ref{sec:conclusion}.

\begin{table}
\centering{}
\caption{
Samples in the train and validation splits across locomotion scores on the left, and the binary setup on the right (1 corresponds to healthy cows and 2 to 5 to lame cows).
}\label{tab:t2}
\vskip -0.1in
\resizebox{0.45\textwidth}{!}{%
\begin{tabular}{lcccccc|cc}
\toprule 
\multicolumn{2}{c}{ } & \multicolumn{5}{c}{Locomotion score} & \multicolumn{2}{c}{Binary setup} \tabularnewline
\midrule 
{ } & {Total} & 1 & 2 & 3  & 4  & 5 & Healthy & Lame \tabularnewline
\midrule 
Validation  & 136 & 80 & 56 & 0 & 0 & 0 & 80 & 56 \tabularnewline
Training & 728 & 523 & 183 & 17 & 3 & 2 & 523 & 205 \tabularnewline
\bottomrule
\end{tabular}
}
\vskip -0.15in
\end{table}

\subsection{Segmentation stage}

\begin{figure}
\centering{}%
\includegraphics[width=0.43\textwidth]{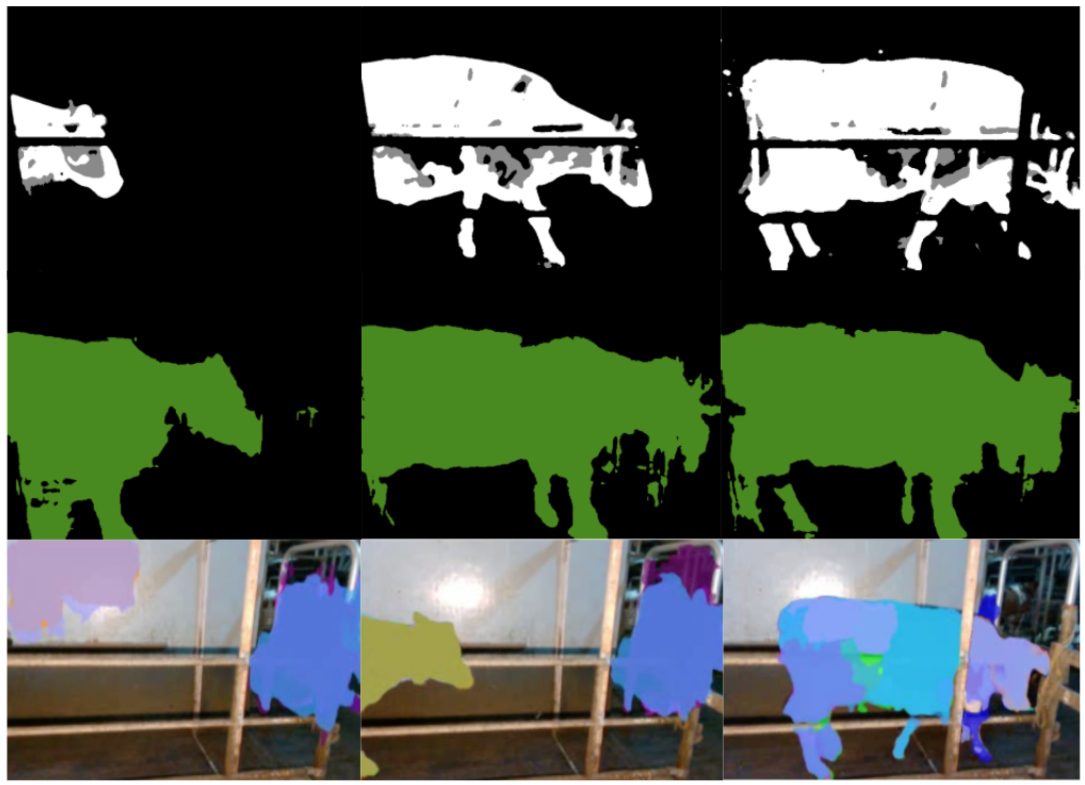}
\vskip -0.1in
\caption{\label{fig:f4} Example of the predicted masks of three off-the-shelf segmentation models, from top to bottom: KNN based background extractor, DeepLabv3~\cite{2017_CVPR_deeplabv3}, and RVOS~\cite{2019_CVPR_RVOS}.}
\vskip -0.2in
\end{figure}

Binary segmentation masks,~i.e. single-channel images where each pixel value indicates the presence or absence of a given object (cows in this case), are a promising solution to highlight the visual features that most influence the assessment of lameness in cows: spine shape and leg distances. 
We experimentally observe that off-the-shelf segmentation models provide low-quality
masks, which fail to highlight the spine or legs of the cow. Figure~\ref{fig:f4} provides sample segmentation masks obtained with three off-the-shelf models; the masks often contain considerable holes in the segmentations, the legs are not detected properly, or the inaccurate contour makes the identification of the spine impossible.

\begin{figure}
\centering{}%

\includegraphics[width=0.43\textwidth]{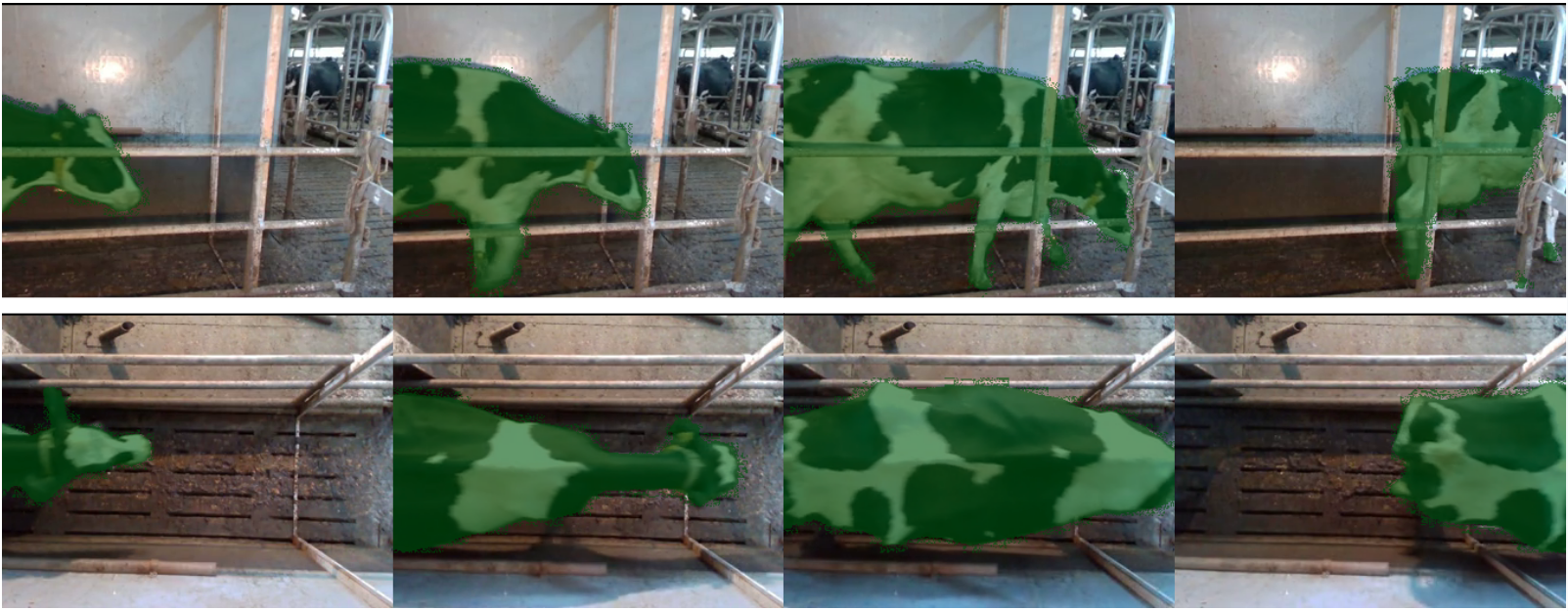}
\vskip -0.1in
\caption{\label{fig:f5} Examples of segmentation masks predicted by the retrained FPN model overlaid in green on the RGB videos.}
\vskip -0.1in
\end{figure}

The main reason for the poor performance of pre-trained segmentation models is the domain shift between the training dataset and the Nedap dataset: while some of these models have been trained in datasets that do contain cows, factors such as changes in the setup of the camera or illumination conditions,
reduce the ability of these models to generalize to new data. To account for the domain shift we re-train a pre-trained model on our dataset. To do this, we manually annotated two sets of 250 frames, each randomly selected across the dataset, for the top and side videos; we ensure that all frames contained a cow; and used the hasty.ai online AI-assisted annotation tool to annotate the frames. See Figure~\ref{fig:f5} (top two rows) for sample segmentation masks obtained through this semi-automatic labeling process.


For the final segmentation model, we choose the FPN model~\cite{2017_CVPR_pyramid} with a ResNeXT~\cite{2017_CVPR_resnext} network as backbone pre-trained on ImageNet~\cite{2015_IJCV_imagenet}. We discard the weights in the decoder of the model and, for each view, top and side, fine-tune the full model end-to-end with our semi-automatically labeled segmentation dataset. The Adam optimizer with a learning rate of $10^{-4}$ was used during training to optimize the dice loss. The model was fine-tuned for 40 epochs using a batch size of 8 and retaining the model with the best validation intersection over union score (IoU). We also used extensive data augmentation during training. Figure~\ref{fig:f5} presents sample outputs of the re-trained model that reaches a validation IoU of 0.95, indicating excellent performance. 

\begin{figure}
\centering{}%
\includegraphics[width=0.43\textwidth]{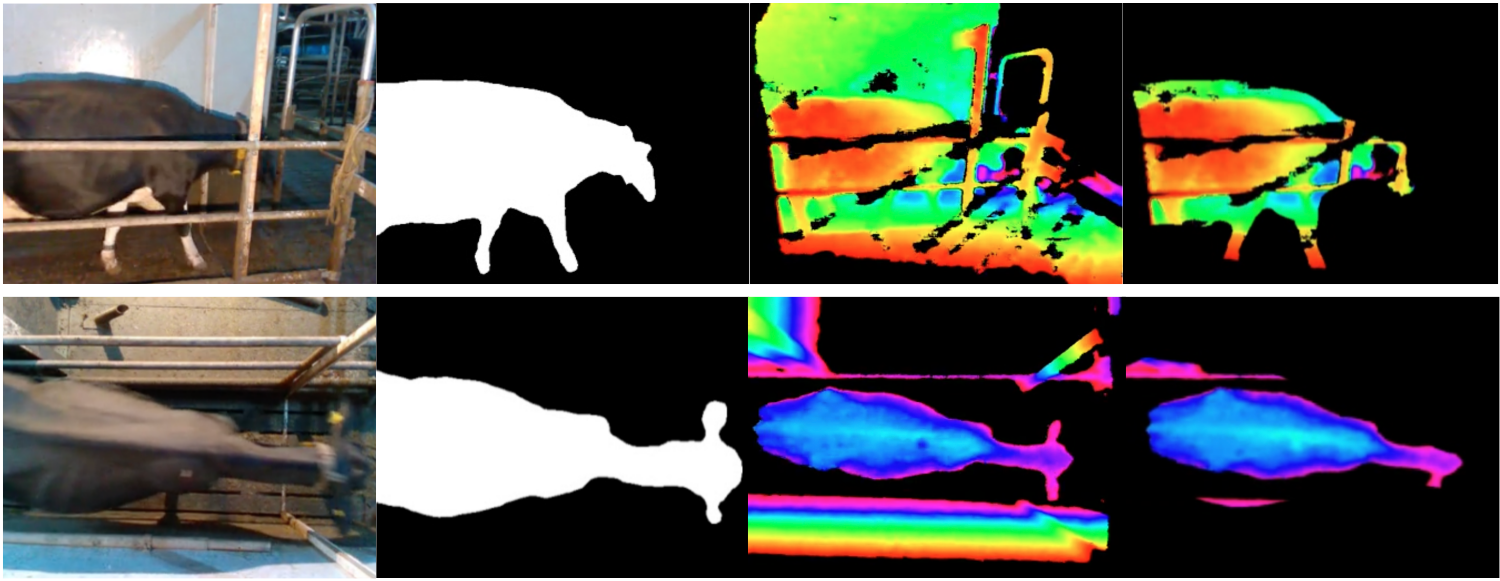}
\vskip -0.1in
\caption{\label{fig:f7} Sample frames of the different videos used as inputs for the model. From left to right: RGB, segmentation masks, HUE-encoded depth maps, and segmentation masked depth maps.}
\vskip -0.2in
\end{figure}

The re-trained model generates high-quality segmentation masks for all the videos in the dataset (both side and top views) that we use as inputs for the feature extractor to encourage the classifier to take into consideration spine curvature and distances between legs. Additionally, we explore the possibility of highlighting these features in the depth videos. To do this we use the predicted segmentation masks from the RGB videos to mask the depth videos resulting in videos that contain depth information only on the regions where the cow has been detected by the segmentation model. We use the resulting videos as input for the feature extractor.  See Figure~\ref{fig:f7} for an example of the different inputs for the different experiments. Note that the rest of the pipeline (video preprocessing and classifier training) remains the same.

\begin{figure*}
\centering{}%
\includegraphics[width=0.93\textwidth]{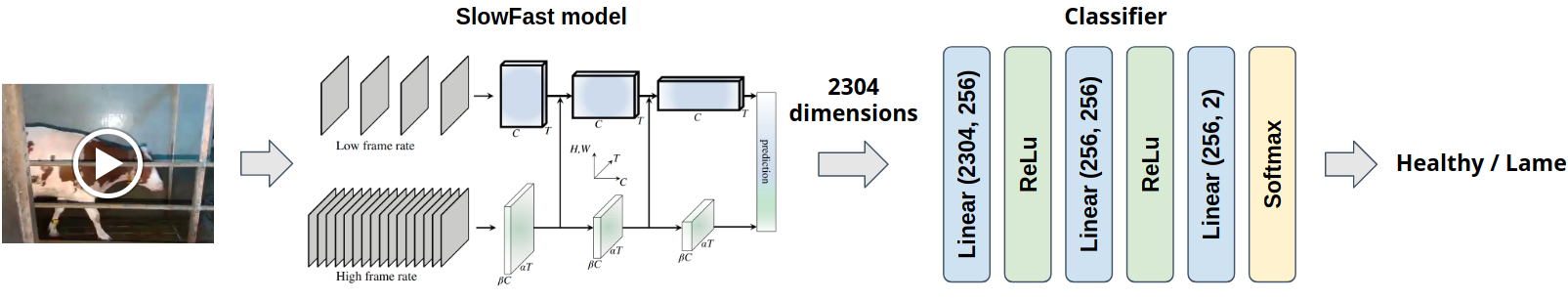}
\vskip -0.1in
\caption{\label{fig:f2} Schematic of the proposed model. The SlowFast model (left) extracts features from the videos and these are used to train the classifier to detect lameness. The multi-layer perceptron (right) used as a classifier receives the extracted features and predicts lameness.}
\vskip -0.15in
\end{figure*}

\subsection{Model}
The proposed model consists of a pre-trained feature extractor that encodes each video in a single feature and a binary classifier that provides a prediction per video: ``healthy" or ``lame." Figure~\ref{fig:f2} provides a schematic of the pipeline. 

\subsubsection{Feature extraction}

We use a SlowFast network~\cite{2019_ICCV_slowfast} to encode the motion and semantics of each video. This model is constituted by two pathways that process each video at two different frame rates: a slow frame rate to capture spatial semantics and a fast frame rate to capture motion at fine temporal resolution. The particular model used 
has a ResNet50~\cite{2016_CVPR_resnet} backbone pre-trained on Kinetics-400~\cite{2017_arxiv_kinetics} using eight frames per video for the slow pathway and 64 for the fast pathway. 

The videos are resized to 340 $\times$ 256 and the frame rate is halved. The features extracted are of 400 and 1904 dimensions for the fast and slow pathways. We empirically observed that concatenating these two features, i.e.~a feature of 2304 dimensions, works better than using either of them individually. Hence, we use these features in all the experiments reported in the remaining of the paper. 

\subsubsection{Training the lameness detector}

To obtain the predictions for each of the videos, we use SlowFast features as the classifier input. The output of the model predicts probabilities of ``healthy" and ``lame." Figure~\ref{fig:f2} summarizes the structure of the classifier: three linear layers followed by a softmax normalization layer. The model was trained for 100 epochs with the Adam~\cite{2014_ICLR_Adam} optimizer using the default PyTorch~\cite{2019_NeurIPS_pytorch} configuration, a learning rate of 0.001, and a batch size of 20. The results of the best model in terms of validation accuracy are reported in Section~\ref{sec:results}.

\section{Results}
\label{sec:results}

Table~\ref{tab:t7} (top) compares the accuracy for the model trained with the four different inputs for the side and the top view. The results under ``RGB" show that the proposed approach gives reasonable performance despite its simplicity. Moreover, the results from training on the segmentation masks, under ``Mask," show an improvement from 61.76\% to 84.56\% when compared to training with the side-view RGB videos. No improvement is observed in the top view videos in this case. Similarly, the results under ``SegmOverDepth" show the effect of applying the segmentation masks over the depth videos, under ``Depth." These experiments also show an improvement in the side view videos from 63.23\% to 75.00\%. However, in this case the performance of the top view videos decreases when masking the depth videos.

Table~\ref{tab:t7} (middle/bottom) provide the recall and precision for the ``lame" class. The results also show an overall improvement when forcing the model to focus on the cow structure through the segmentation masks. Unlike the accuracy results, the recall improves in both top and side views
when using the segmentation masks instead of the RGB videos. Similarly, the precision also improves for both views when using the segmentation masks over the depth videos.


\begin{table}
\centering{}
\caption{Accuracy metrics for model predictions.}\label{tab:t7}
\vskip -0.1in
\resizebox{0.45\textwidth}{!}{%
\begin{tabular}{llcccc}
\toprule 
& { } & {RGB} & {Mask} & {Depth} & {SegmOverDepth} \tabularnewline
\midrule 
Accuracy (\%) & {Top} & 65.44 & 65.44 & \textbf{76.47} & 75.00 \tabularnewline
& {Side} & 61.76 & \textbf{84.56} & 63.23 & 75.00  \tabularnewline
\midrule
Precision & {Top} & 0.67 & 0.64 & 0.73 & \textbf{0.76} \tabularnewline
& {Side} & 0.58 & \textbf{0.82} & 0.62 & 0.71 \tabularnewline
\midrule
Recall & {Top} & 0.32 & 0.36 & \textbf{0.68} & 0.57 \tabularnewline
& {Side} & 0.27 & \textbf{0.80} & 0.27 & 0.66 \tabularnewline
\bottomrule
\end{tabular}
}
\vskip -0.15in
\end{table}

\section{Conclusion}
\label{sec:conclusion}
This paper investigated three important design questions for using computer vision to predict lameness in cows: 1) the impact of camera position, 2) the impact of features used for classification, and 3) the classes used for classification. Using a proprietary dataset and a state-of-the-art classifier, we studied the impact of answers to these design questions on the lameness classification performance. 

For the camera position, we found that recordings from the side give stronger performance than from the top ($0.85$ and $0.76$ in terms of accuracy). We consider the lower results from the top view as a limitation to be addressed. This camera position is more practical since it does not require a dedicated space on the side of a corridor.

For the input data used in the classification, we investigated full-frame RGB data, full-frame depth data encoded as hue colors, and corresponding masked versions using only the automatically segmented cows in the video. For recording from the side, RGB data produced stronger performance than depth data, while for the top view this was reversed. For both, RGB and depth data, the masking improved the performance. This shows that segmentation masks force the feature extractor to leverage relevant characteristics of the cow structure (spine curvature and leg distances) resulting in better lameness classification. 

As most herds have more healthy animals than lame, the distribution of locomotion scores in our dataset was skewed, making the severe cases hard to predict. We combined the four scores indicating degrees of lameness into a class ``lame." This improved class balance while losing information about the degree of lameness. We leave addressing this imbalance in a more principled manner for future work. 

Finally, from a more practical side, the preprocessing of the depth maps into HUE-encoded images is a computationally expensive process that should be addressed before deploying the system on farms.

\section*{Acknowledgment}

We acknowlege Roxie Muller and Arnold Harbers of Nedap for their advisory role and help in setting up the data collection. This publication has emanated from research conducted with the financial support of Science Foundation Ireland (SFI) under grant number SFI/15/SIRG/3283 and SFI/12/RC/2289\_P2.

{\small
\bibliographystyle{ieee_fullname}
\bibliography{egbib}

\begin{thebibliography}{10}\itemsep=-1pt

\bibitem{2021_ICCVw_grass}
Paul Albert, Mohamed Saadeldin, Badri Narayanan, Brian Mac~Namee, Deirdre
  Hennessy, Aisling O'Connor, Noel O'Connor, and Kevin McGuinness.
\newblock Semi-supervised dry herbage mass estimation using automatic data and
  synthetic images.
\newblock In {\em IEEE/CVF International Conference on Computer Vision workshop
  (ICCVw)}, 2021.

\bibitem{2019_elsevier_review}
Maher Alsaaod, Mahmoud Fadul, and Adrian Steiner.
\newblock Automatic lameness detection in cattle.
\newblock {\em The veterinary journal}, 2019.

\bibitem{2017_CVPR_deeplabv3}
Liang-Chieh Chen, George Papandreou, Florian Schroff, and Hartwig Adam.
\newblock Rethinking atrous convolution for semantic image segmentation.
\newblock In {\em IEEE/CVF Conference on Computer Vision and Pattern
  Recognition (CVPR)}, 2017.

\bibitem{2022_WACVw_tempConst}
Julia Dietlmeier, Feiyan Hu, Frances Ryan, Noel~E O'Connor, and Kevin
  McGuinness.
\newblock Improving person re-identification with temporal constraints.
\newblock In {\em IEEE/CVF Winter Conference on Applications of Computer Vision
  workshop (WACVw)}, 2022.

\bibitem{2021_CVPR_transfer}
Linus Ericsson, Henry Gouk, and Timothy~M Hospedales.
\newblock How well do self-supervised models transfer?
\newblock In {\em IEEE/CVF Conference on Computer Vision and Pattern
  Recognition (CVPR)}, 2021.

\bibitem{2019_ICCV_slowfast}
Christoph Feichtenhofer, Haoqi Fan, Jitendra Malik, and Kaiming He.
\newblock Slowfast networks for video recognition.
\newblock In {\em IEEE/CVF international conference on computer vision (ICCV)},
  2019.

\bibitem{2016_CVPR_resnet}
Kaiming He, Xiangyu Zhang, Shaoqing Ren, and Jian Sun.
\newblock Deep residual learning for image recognition.
\newblock In {\em IEEE conference on computer vision and pattern recognition
  (CVPR)}, 2016.

\bibitem{2021_sensors_review}
Xi Kang, Xu~Dong Zhang, and Gang Liu.
\newblock A review: Development of computer vision-based lameness detection for
  dairy cows and discussion of the practical applications.
\newblock {\em Sensors}, 2021.

\bibitem{2021_AAAI_dl4lameness}
Yasmine Karoui, Amanda A~Boatswain Jacques, Abdoulaye~Banir{\'e} Diallo, Elise
  Shepley, and Elsa Vasseur.
\newblock A deep learning framework for improving lameness identification in
  dairy cattle.
\newblock In {\em Proceedings of the AAAI Conference on Artificial
  Intelligence}, 2021.

\bibitem{2017_arxiv_kinetics}
Will Kay, Joao Carreira, Karen Simonyan, Brian Zhang, Chloe Hillier, Sudheendra
  Vijayanarasimhan, Fabio Viola, Tim Green, Trevor Back, Paul Natsev, et~al.
\newblock The kinetics human action video dataset.
\newblock {\em arXiv:1705.06950}, 2017.

\bibitem{2014_ICLR_Adam}
Diederik~P Kingma and Jimmy Ba.
\newblock Adam: A method for stochastic optimization.
\newblock In {\em International Conference on Learning Representations (ICLR)},
  2014.

\bibitem{2021_MDPI_review}
Guoming Li, Yanbo Huang, Zhiqian Chen, Gary~D Chesser, Joseph~L Purswell, John
  Linhoss, and Yang Zhao.
\newblock Practices and applications of convolutional neural network-based
  computer vision systems in animal farming: A review.
\newblock {\em Sensors}, 2021.

\bibitem{2017_CVPR_pyramid}
Tsung-Yi Lin, Piotr Doll{\'a}r, Ross Girshick, Kaiming He, Bharath Hariharan,
  and Serge Belongie.
\newblock Feature pyramid networks for object detection.
\newblock In {\em IEEE conference on computer vision and pattern recognition
  (CVPR)}, 2017.

\bibitem{2020_elsevier_cowStructure}
He Liu, Amy~R Reibman, and Jacquelyn~P Boerman.
\newblock Video analytic system for detecting cow structure.
\newblock {\em Computers and Electronics in Agriculture}, 2020.

\bibitem{2019_NeurIPS_pytorch}
Adam Paszke, Sam Gross, Francisco Massa, Adam Lerer, James Bradbury, Gregory
  Chanan, Trevor Killeen, Zeming Lin, Natalia Gimelshein, Luca Antiga, et~al.
\newblock {PyTorch}: An imperative style, high-performance deep learning
  library.
\newblock {\em Advances in neural information processing systems (NeurIPS)},
  2019.

\bibitem{2018_arxiv_yolov3}
Joseph Redmon and Ali Farhadi.
\newblock Yolov3: An incremental improvement.
\newblock {\em arXiv:1804.02767}, 2018.

\bibitem{2015_IJCV_imagenet}
Olga Russakovsky, Jia Deng, Hao Su, Jonathan Krause, Sanjeev Satheesh, Sean Ma,
  Zhiheng Huang, Andrej Karpathy, Aditya Khosla, Michael Bernstein, et~al.
\newblock {ImageNet} large scale visual recognition challenge.
\newblock {\em International journal of computer vision}, 2015.

\bibitem{image_compression}
Tetsuri Sonoda and Anders Grunnet-Jepsen.
\newblock Depth image compression by colorization for {Intel}® {RealSense}™
  depth cameras.

\bibitem{2019_CVPR_RVOS}
Carles Ventura, Miriam Bellver, Andreu Girbau, Amaia Salvador, Ferran Marques,
  and Xavier Giro-i Nieto.
\newblock Rvos: End-to-end recurrent network for video object segmentation.
\newblock In {\em IEEE/CVF Conference on Computer Vision and Pattern
  Recognition (CVPR)}, 2019.

\bibitem{2021_elsevier_eeg}
Zitong Wan, Rui Yang, Mengjie Huang, Nianyin Zeng, and Xiaohui Liu.
\newblock A review on transfer learning in {EEG} signal analysis.
\newblock {\em Neurocomputing}, 2021.

\bibitem{2021_Wiley_review}
Tonghe Wang, Yang Lei, Yabo Fu, Jacob~F Wynne, Walter~J Curran, Tian Liu, and
  Xiaofeng Yang.
\newblock A review on medical imaging synthesis using deep learning and its
  clinical applications.
\newblock {\em Journal of applied clinical medical physics}, 2021.

\bibitem{2020_elsevier_YOLOlameness}
Dihua Wu, Qian Wu, Xuqiang Yin, Bo Jiang, Han Wang, Dongjian He, and Huaibo
  Song.
\newblock Lameness detection of dairy cows based on the yolov3 deep learning
  algorithm and a relative step size characteristic vector.
\newblock {\em Biosystems Engineering}, 2020.

\bibitem{2017_CVPR_resnext}
Saining Xie, Ross Girshick, Piotr Doll{\'a}r, Zhuowen Tu, and Kaiming He.
\newblock Aggregated residual transformations for deep neural networks.
\newblock In {\em IEEE conference on computer vision and pattern recognition
  (CVPR)}, 2017.

\end{thebibliography}
}

\end{document}